\documentclass[letterpaper, 10pt, conference]{ieeeconf}
\IEEEoverridecommandlockouts
\usepackage{cite}
\usepackage{amsmath,amssymb,amsfonts}
\usepackage{algorithmic}
\usepackage{graphicx}
\usepackage{textcomp}
\usepackage{xcolor}
\usepackage{dsfont} 
\usepackage{booktabs}
\let\labelindent\relax
\usepackage{enumitem}
\usepackage[draft]{hyperref}
\usepackage{hyperref}
\def\BibTeX{{\rm B\kern-.05em{\sc i\kern-.025em b}\kern-.08em
    T\kern-.1667em\lower.7ex\hbox{E}\kern-.125emX}}
\begin{document}

\title{\LARGE \bf Learning Vision-Based Agile Gap Traversal: Differentiable Simulation with a Warm-Started Critic
}

\author{
    Nuthasith Gerdpratoom, Tianchen Sun, Yichao Gao, and Lin Zhao
    \thanks{
        The authors are with the Department of Electrical and Computer Engineering, National University of Singapore, 4 Engineering Drive 3, Singapore 117583, Singapore. Email: 
        {\tt\small 
            \{nuthasith, tianchen.sun, yichao\_gao\}@u.nus.edu, zhaolin@nus.edu.sg
        }
    }
}



\maketitle

\begin{abstract}

Traversing narrow gaps is challenging for autonomous quadrotors, especially when control commands come directly from high-dimensional visual observations. Existing end-to-end methods often rely on behavior cloning or full-rollout backpropagation through time (BPTT) via differentiable simulation, which can limit policy performance or incur high training costs. We propose a two-stage reinforcement learning framework for more efficient ego-centric visuomotor gap-traversal policy training, leveraging quasi-analytical policy gradients (QPG) via differentiable simulation and critic warm-starting. The framework utilizes QPG to avoid backpropagation through visual rendering, reducing computation and memory costs while improving sample efficiency. In the first stage, an expert actor and critic are trained using privileged observations, including gap geometry. Unlike prior gap-traversal approaches, our training utilizing QPG does not require resetting the agent along optimized reference trajectories. In the second stage, a visual policy is trained using binary gap masks from two ego-centric cameras and low-dimensional observations, while its privileged critic is warm-started from the first stage. This substantially improves training efficiency and traversal success compared with cold-starting the critic or using full-rollout BPTT. Our framework does not require retraining the expert actor when system parameters change, enabling more efficient generalization across drone platforms than state-of-the-art visual gap-traversal methods based on action supervision. The learned visual policy also generalizes to gaps with unseen shapes. Extensive real-world experiments further demonstrate robust gap traversal using binary masks rendered online. Beyond gap traversal, the proposed framework is generic and can be extended to other visuomotor robot learning tasks.

\end{abstract}


\section{Introduction}

Maneuvering through tight gaps is one of the most challenging tasks in quadrotor autonomous navigation in obstructed environments, especially when the gaps are much narrower than the vehicle's lateral dimension. In such cases, the agent must adjust its attitude to pass through narrow gaps, requiring robust $SE(3)$ planning and control \cite{lee2010geometric, mellinger2011minimum, falanga2017aggressive, liu2018search}.
Traditional autonomous navigation is usually structured as a modular pipeline, which propagates and compounds errors throughout the system.
This has motivated deep learning approaches in which neural network policies are trained for gap traversal~\cite{xiao2021flying, chen2022learning, lin2019flying, xie2023learning, hu2025narrow, zhang2026vision}, enabling the direct use of high-dimensional observations, such as images, as policy inputs.




\begin{figure}
    \centering
    \includegraphics[width=1\linewidth]{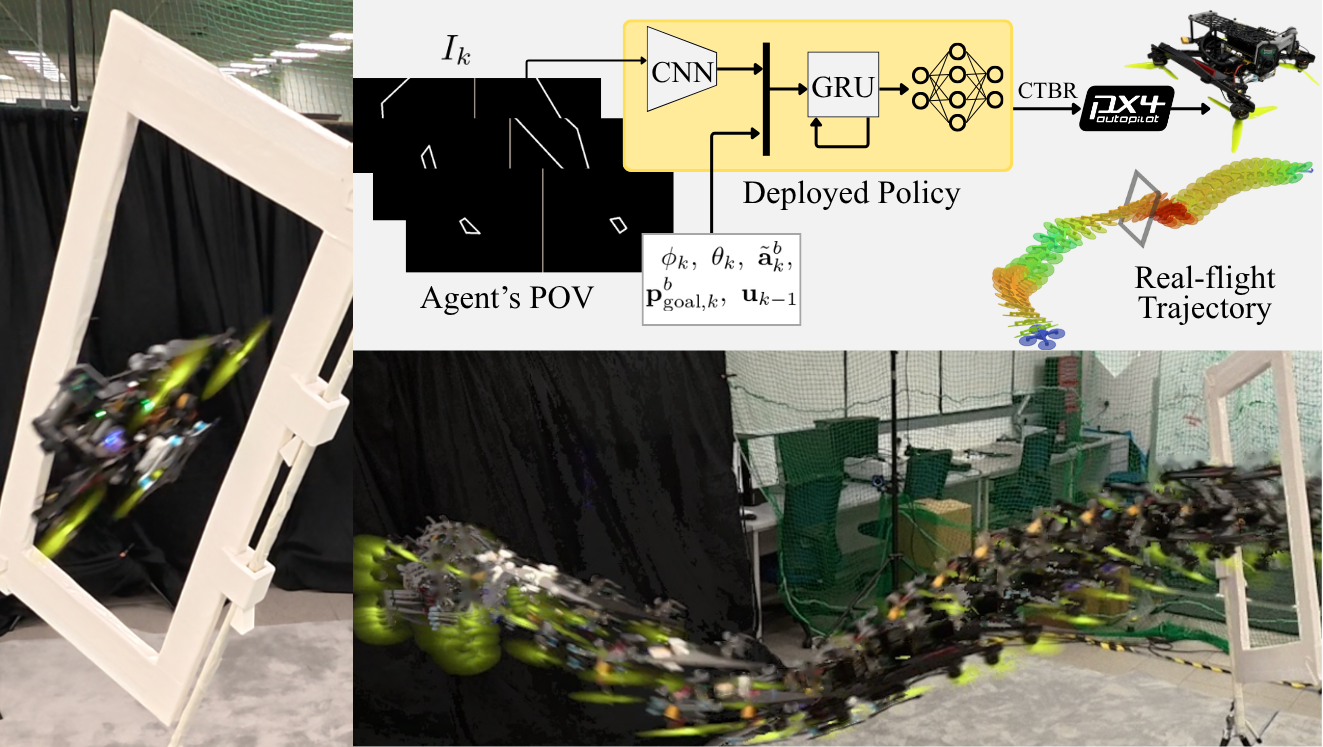}
    \caption{Visual agile gap traversal. Top right: the deployed policy maps the binary gate mask $I_k$ from ego-centric cameras, along with low-dimensional observations (the agent's roll $\phi_k$, pitch $\theta_k$, accelerometer reading $\tilde{\mathbf{a}}^b_k$, goal position $\mathbf{p}_{\text{goal},k} ^b$, and past action $\mathbf{u}_{k-1}$) to thrust and body-rate commands. Left and bottom right: a quadrotor executes the policy in real flight.}
    \label{fig:placeholder}
    \vspace{-0.6cm}
\end{figure}


In particular, model-free reinforcement learning (RL) approaches generally rely on estimated policy gradients while treating the system dynamics as a black box. However, these estimates can exhibit high variance, causing unstable training and sample inefficiency. This motivates utilizing analytical policy gradients (APG) for controller learning \cite{wiedemann2023training}. APG uses a differentiable simulator, allowing gradients to flow through trajectories and yielding low-variance policy gradients, thereby improving sample efficiency. However, it suffers from vanishing or exploding gradients and getting stuck in local minima for long-horizon tasks. Short-Horizon Actor-Critic (SHAC) \cite{xu2022accelerated} circumvents these problems. Instead of performing BPTT over an entire episode, SHAC truncates the computational graph into shorter segments and incorporates a value function, providing gradients while accounting for long-term returns.
\cite{you2026accelerating} extends SHAC to a partially observable control problem, where the observation is excluded from the gradient computation, avoiding large gradients from pixels. The policy is then updated via quasi-analytical policy gradient (QPG), which provides both long-term credit and an implicit feedback pathway from the smooth learned critic rather than from the observations.

Our setup of the gap traversal, shown in Fig. \ref{fig:dimensions}, leaves $6$ cm of lateral clearance per side for a perfectly aligned traversal and allows only $23^\circ$ of roll misalignment. Relying on ego-centric vision sensing also poses significant challenges, as the agent must infer the gap geometry from high-dimensional visual cues, which are tightly coupled with its agile movement. \cite{wu2025whole} and \cite{wu2026precise} demonstrate distilling an expert policy, trained via RL with the known gap geometry, into a vision-based policy that accepts a binary mask. The policy is supervised to reproduce the expert's actions that are optimal only for the training plant and, thus, cannot adapt when the dynamics shift. Moreover, training the expert policy requires trajectory optimization to avoid undirected exploration, which adds computational overhead to the pipeline. \cite{zhang2026vision} proposes policy optimization via a differentiable simulator with depth observation, showing the potential of first-order policy gradients for this task. Unlike depth images, binary masks provide shapes without direct ranges, demanding that the policy infer geometry from them. In this work, we directly use the binary masks of the gap to reduce the dependence on visual textures and to support vectorized rendering alongside the physics simulator. A binary mask encoder can be further deployed to achieve a visuomotor policy from the RGB image. Nevertheless, the derivatives of the binary masks are zero almost everywhere. QPG decouples observations from backpropagation and propagates gradients through short rollouts, with a privileged critic estimating the credit beyond the horizon \cite{xu2022accelerated, you2026accelerating}. Full-episode BPTT in the previous work propagates gradients through all steps, which can cause exploding gradients and hinder optimization. Although QPG is effective, it depends more on the critic than SHAC does. A randomly initialized critic offers no guidance until it has been fitted on the visual actor's rollouts, which is poor at first. Since the critic sees the privileged state, it can warm-start another actor. We therefore pretrain a privileged expert and its critic, then warm-start that critic to train the visual actor. For this task, this improves performance over training an actor-critic from scratch with QPG.

We propose a two-stage training framework for vision-based agile gap traversal via differentiable simulation. In the first stage, an expert policy and critic are trained jointly via QPG. In the second stage, the pretrained critic warm-starts the value function for learning a visual policy, which observes binary gate masks from two ego-centric cameras and low-dimensional inputs.
Our contributions are threefold. First, we introduce an efficient two-stage reinforcement learning framework that leverages short-horizon quasi-analytical policy gradient (QPG) and privileged critic pretraining; the warm-started critic accelerates and stabilizes visuomotor policy learning in the second stage (\ref{sec:vision_policy}).  To the best of our knowledge, we are the first to apply short-horizon QPG to visual agile narrow-gap traversal, and the framework proposed is not limited to visual agile gap traversal and can extend to other challenging robotics tasks. Second, compared with the state-of-the-art reinforcement-imitation learning pipeline for gap traversal \cite{wu2025whole, wu2026precise}, our design eliminates the need to reset agents along optimized trajectories during expert training and action supervision (\ref{sec:expert}). Moreover, compared with policy distillation, our method does not require retraining the expert actor or critic for plant-parameter changes, improving generalization across drone platforms (\ref{subsec:plant_revise}). Third, our trained policy generalizes to gaps with previously unseen shapes, and we further demonstrate real-world vision-based quadrotor narrow-gap traversal in confined spaces with binary masks rendered online.

\section{Related Work}

\subsection{Agile Gap Traversal for Quadrotors}
In agile gap traversal, traditional approaches plan a dynamically feasible $SE(3)$ trajectory from the gap geometry, then track it using a lower-level policy \cite{falanga2017aggressive, liu2018search}. Data-driven techniques learn the policy via expert demonstration or agent-environment interaction. Given the gap pose, \cite{lin2019flying} imitates a model-based pipeline and fine-tunes with RL, \cite{xiao2021flying} trains an end-to-end policy via model-free RL with a curriculum, \cite{chen2022learning} incorporates safety-aware exploration, and \cite{wang2023learning, romero2025actor, sun2026learning} train a network to adaptively parameterize a predictive controller. With onboard sensing, \cite{xie2023learning} detects the gap from a camera for RL training, \cite{hu2025narrow} and \cite{zhang2026vision} map depth images to commands via differentiable physics, and \cite{wu2025whole} and \cite{wu2026precise} distill a model-free RL-trained expert, which observes the gap edges and is reset along optimal trajectories, into a pixel-based student. Our policy observes binary gate masks from two ego-centric cameras, learns via first-order gradients, and requires neither a trajectory optimizer nor distillation. We exploit the actor-critic structure by feeding exposed privileged information into the critic network during training, while the actor remains identical to the deployed policy \cite{pinto2018asymmetric, Geles-RSS-24}.

\subsection{Policy Optimization via Differentiable Simulator}
Policy optimization via a differentiable simulator has gained attention. Sample-based policy optimization algorithms \cite{schulman2015trust, schulman2017proximal, haarnoja2018soft} require a large number of agent-environment interactions to estimate gradients. APG via a differentiable simulator provides low-variance gradients and thus more sample-efficient learning \cite{freeman2021brax, wiedemann2023training}, although BPTT over long horizons yields chaotic gradients, which motivates the short-horizon and decoupled formulations \cite{xu2022accelerated, you2026accelerating} that we build on. For quadrotors, differentiable physics has been used to learn recovery from visual features \cite{heeg2025learning}, agile navigation using a depth camera \cite{zhang2025learning}, obstacle avoidance from optical flow \cite{hu2025seeing}, drone racing with vector-field-augmented gradients \cite{su2026vector}, and gap traversal from depth images \cite{hu2025narrow, zhang2026vision}.
Most quadrotor work optimizes the policy via BPTT through rollouts without a value function, which requires far more samples to achieve a comparable success rate. Instead, we adopt QPG with a privileged critic and backpropagate through a simplified model. Hence, the renderer is detached from the gradient path, and the observation can be chosen for the task rather than for differentiability. We also use binary masks as visual cues instead of depth images, which omit the gap's geometry, posing an additional challenge for policy learning.

\section{Methodology}



\subsection{Problem Formulation}

\begin{figure}
    \vspace*{2mm}
    \centering
    \includegraphics[width=1\linewidth]{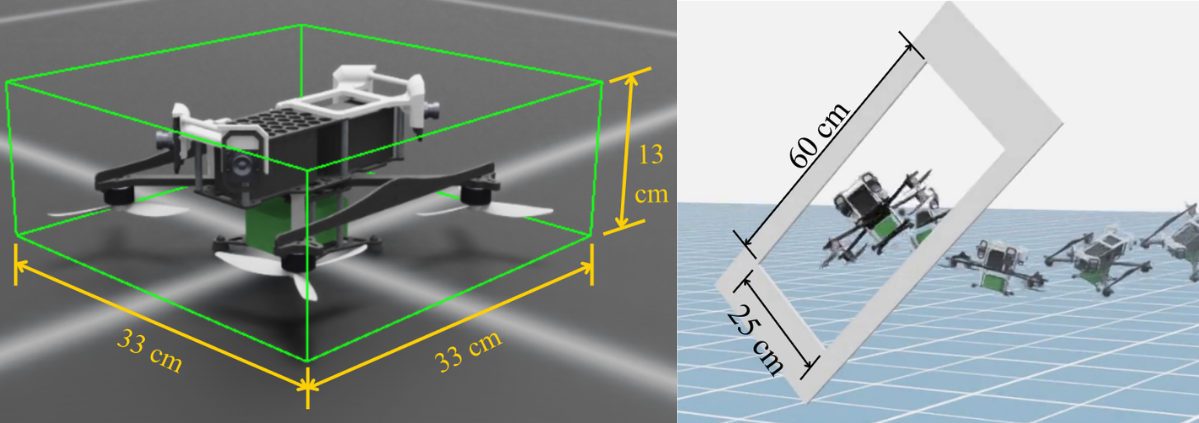}
    \caption{Left: the bounding box (transparent green) $\mathcal{C}$, representing the quadrotor's collision. Right: opening dimensions of the gap $\mathcal{G}$. Both are visualized through Isaac Sim.}
    \label{fig:dimensions}
    \vspace{-0.4cm}
\end{figure}
\begin{figure*}
    \vspace*{2mm}
    \centering
    \includegraphics[width=\textwidth]{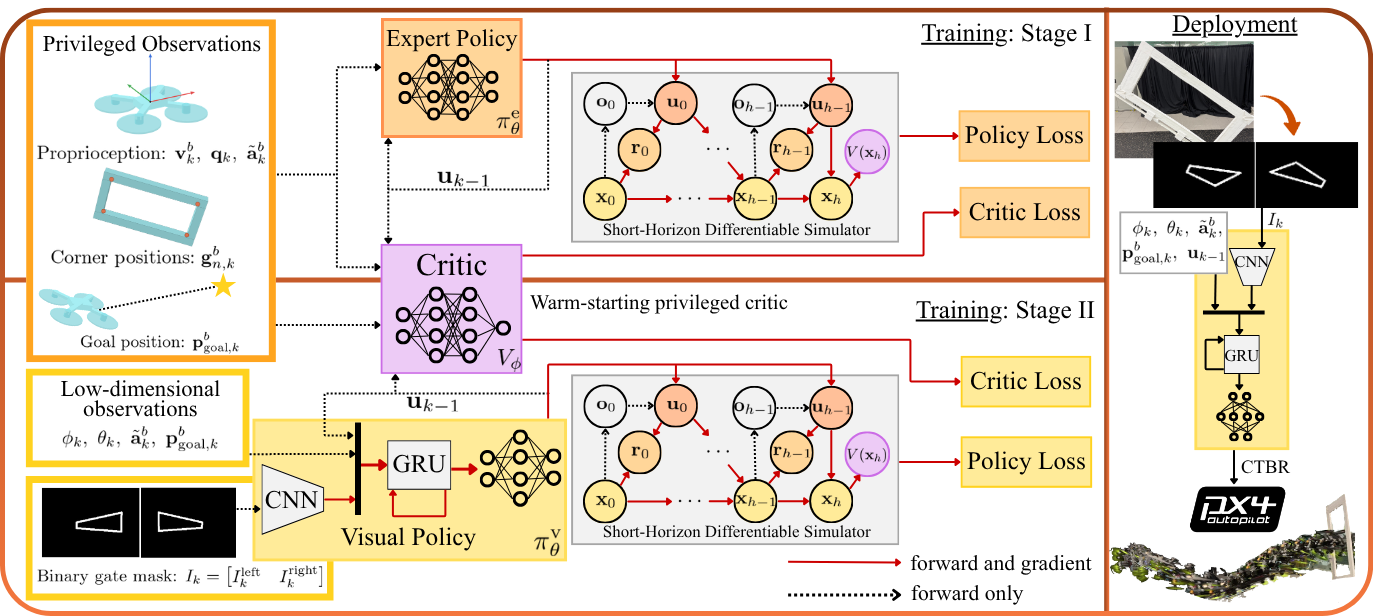}
    \caption{Overview of the proposed two-stage training pipeline. In the first stage, a critic $V_\phi$ and an expert policy $\pi_\theta^{\text{e}}$ are trained through QPG in a symmetric actor-critic manner, where the networks accept privileged observations: states (agent's velocity in its body frame $\mathbf{v}^b_k$, quaternion  $\mathbf{q}_k,$ and accelerometer readings $\tilde{\mathbf{a}}_k^b$), corner positions $\mathbf{g}^b_{n,k}$ for $n \in [0,3] \cap \mathbb{Z}$, goal position in body frame $\mathbf{p}^b_{\text{goal}, k}$, and previous action $\mathbf{u}_{k-1}$. In the second stage, the critic is copied from the pretraining phase to warm-start the value function. The policy is trained in an asymmetric actor-critic fashion, accepting a binary gate mask from the two ego-centric cameras and low-dimensional observations. The right side shows the policy execution during deployment.}
    \label{fig:method}
    \vspace{-0.4cm}
\end{figure*}

The task is for a quadrotor to traverse a tight gap agilely without colliding with obstacles and then hover stably at a goal position. The quadrotor is modeled as a rigid body with the state evolution defined as $\mathbf{x}_{i+1} = f(\mathbf{x}_i, \mathbf{u}_i)$, where $\mathbf{x}_i \in \mathcal{X} \subset \mathbb{R}^{n}$ and $\mathbf{u}_i \in \mathcal{U} \subset \mathbb{R}^m$ are the system's state and control input at time $i \in \mathbb{Z}$, respectively. We formalize the gap as a slab shape with a rectangular opening. The geometric collision of the quadrotor is defined as an oriented bounding box $\mathcal{C}(\mathbf{x}_i)$, and the gap is described as a thin planar gate frame  $\mathcal{G}$ with the inner opening. The dimensions are specified in Fig. \ref{fig:dimensions}. The drone traverses the gate cleanly if and only if the centroid of the bounding box passes through the opening and there is no intersection between the box and the gate frame at every time step $i$, i.e., $\mathcal{C}(\mathbf{x}_i) \cap \mathcal{G} = \varnothing, \ \forall i$. The task is considered successful if the drone traverses cleanly and stabilizes at the goal for a period of time thereafter. Each control step contains $N = \Delta t_c/\Delta t_p \in \mathbb{N}$ physics substeps, where $\Delta t_c$ and $\Delta t_p$ are the control and physics intervals in seconds. We denote $i$ as a physics step, and $k = \lfloor i/N \rfloor$ as a control step. A zero-order hold is applied to the input throughout the control step, i.e., $\mathbf{u}_i = \mathbf{u}_k$ for $i \in \big[kN,(k+1)N \big)\cap \mathbb{Z},\ \forall k$. The feedback controller is defined as a stochastic policy $\pi_\theta(\cdot|\mathbf{o}_k)$, modeled as a Gaussian distribution, where $\theta$ is the parameter, and $\mathbf{o}_k \in \mathcal{O}$ is an observation vector, obtained through a sensor model. The policy is expressed through the reparameterization trick as follows to enable learning via backpropagation:
\begin{equation}
    \label{eq:reparameterisation}
    \mathbf{u}_k = \tanh\big(\boldsymbol{\mu}_\theta(\mathbf{o}_k) +\boldsymbol{\epsilon}_k\boldsymbol{\sigma}_\theta \big), \quad \boldsymbol{\epsilon}_k \sim \mathcal{N}(0, \mathbf{I}_m)
\end{equation}
where $\boldsymbol{\mu}_\theta \in \mathbb{R}^m$ and $\boldsymbol{\sigma}_\theta \in \mathbb{R}^m$ are the mean and standard deviation. The deployed policy accepts a visual observation (i.e., the gate's binary mask $I_k$, as shown in Fig. \ref{fig:method} right), and extra low-dimensional observations (i.e., roll $\phi_k$, pitch $\theta_k$, accelerometer reading $\tilde{\mathbf{a}}_k^b$, and goal position in body frame $\mathbf{p}_{\text{goal}, k}^b$), and then outputs collective thrust and body rates (CTBR) commands to an autopilot.




\subsection{Quadrotor Dynamics}

The state evolution of a quadrotor is governed by the continuous-time rigid body dynamics as follows:
\begin{equation}
    \label{eq:translational_dynamics}
    \dot{\mathbf{p}} = \mathbf{v}, \quad
    \dot{\mathbf{v}} = \frac{1}{m}R(\mathbf{q})\mathbf{f}^b - g \hat{\mathbf{z}},
\end{equation}
where $\mathbf{p} \in \mathbb{R}^3$ and $\mathbf{v} \in \mathbb{R}^3$ are position and velocity in world frame. $\mathbf{q} \in \mathbb{R}^4$ is a quaternion from the body frame to the world frame. $R(\mathbf{q}) \in SO(3)$ is a rotation matrix. $\mathbf{f}^b = [ 0,0,T ]^\top$ is a body-frame force acting on the body, and $T \in \mathbb{R}_+$ is thrust. $m \in \mathbb{R}$ is a quadrotor's mass in kg. $g = 9.81$ m/s$^2$ is the acceleration due to the Earth's gravity, multiplied by a unit vector $\hat{\mathbf{z}} = [0,0,1]^{\top}$. For the attitude, $\mathbf{q}$ evolves under the following quaternion kinematics:
\begin{equation}
    \label{eq:attitude_kinemetics}
    \dot{\mathbf{q}} = \frac{1}{2} \mathbf{q} \otimes
        \begin{bmatrix}
        \boldsymbol{\omega }\\
        0
        \end{bmatrix},
\end{equation}
where $\boldsymbol{\omega} \in \mathbb{R}^3$ is an achieved body rate in body frame, and $\otimes$ is a quaternion multiplication operator \cite{chou1992quaternion}.

In this work, we approximate the autopilot's closed-loop body rate dynamics as a second-order low-pass filter with internal states $[\dot{\boldsymbol{\omega}}^\top, \boldsymbol{\omega}^\top ]^\top$, defined as:
\begin{equation}
    \label{eq:inner_body_rate}
    \ddot{\boldsymbol{\omega}} + 2 \zeta \omega_n \dot{\boldsymbol{\omega}} + \omega_n^2 \boldsymbol{\omega} = \omega_n^2 K \boldsymbol{\omega}^{\text{ref}}
\end{equation}
with constant scalar parameters $\omega_n, \zeta,$ and $K$. The achieved thrust $T$ is also modeled as a first-order filter to represent the actuation delay with a time constant $\tau$ as:
\begin{equation}
    \label{eq:inner_thrust}
    \tau \dot{T} + T = T^{\text{ref}}.
\end{equation}


The state equations are discretized with a physics substep interval $\Delta t_p$. $T_i$ and $\boldsymbol{\omega}_i$ are updated first through the exact discretized inner-loop models of \eqref{eq:inner_body_rate} and \eqref{eq:inner_thrust}, respectively, and then are used to integrate the rigid body dynamics as:
\begin{equation}
    \label{eq:rigid_body_dyn_dis}
    \begin{aligned}
        \mathbf{v}_{i+1} &= \mathbf{v}_i + \dot{\mathbf{v}}_i \Delta t_p, \\
        \mathbf{p}_{i+1} &= \mathbf{p}_i + \mathbf{v}_i \Delta t_p + \frac{1}{2} \dot{\mathbf{v}}_i \Delta t_p^2, \\
        \mathbf{q}_{i+1} &= \frac{\mathbf{q}_i+ \dot{\mathbf{q}}_i \Delta t_p}{\| \mathbf{q}_i+ \dot{\mathbf{q}}_i \Delta t_p\|_2}.
    \end{aligned}
\end{equation}
From \eqref{eq:rigid_body_dyn_dis} and the discrete-time inner-loop dynamics models, the quadrotor's state is $\mathbf{x}_i = [\mathbf{p}_i^\top, \mathbf{v}_i^\top, \mathbf{q}_i^\top, \dot{\boldsymbol{\omega}}_i^\top, \boldsymbol{\omega}_i^\top, T_i]^\top \in \mathbb{R}^{17}$, 
$T^{\text{ref}}_k = 0.5 T^{\max} (\mathbf{u}_{k, [0]} +1) \in \mathbb{R}$, and $\boldsymbol{\omega}_k^\text{ref} = \omega^{\max} \mathbf{u}_{k,[1:3]} \in \mathbb{R}^3$ (the last three entries of $\mathbf{u}_k$), for all physics step $i$ and control step $k$, where $T^{\text{max}}$ and $\omega^{\text{max}}$ are the maximum thrust and body rate, respectively.


\subsection{Policy Optimization via Differentiable Simulation}

In this work, policy learning via differentiable physics is based on QPG by minimizing SHAC's actor loss:

\begin{equation}
    \label{eq:actor_loss}
    \mathcal{L}_{\theta} = - \frac{1}{h} \left[\sum_{k=k_0}^{k_0+h-1} \gamma^{k-k_0} \mathcal{R} (\mathbf{x}_k,\mathbf{u}_k(\mathbf{o}_k)) + \gamma^ h V_\phi(\mathbf{c}_{k_0 + h}) \right],
\end{equation}
where $\mathcal{R}(\mathbf{x}_k, \mathbf{u}_k)$ is a single-step total reward, $\gamma \in (0,1)$ is a discount factor, a short horizon $h \in \mathbb{N}$, and privileged observations $\mathbf{c}_k = \Psi(\mathbf{x_k}, \mathcal{G})$.
A rollout consists of the sensor model, the policy with recurrent state $\mathbf{h}_k$ (which exists only in the second stage), and the dynamics $f$. The total derivative of the action can be derived as:
\[
    \frac{d \mathbf{u}_k}{d \theta } = \frac{\partial \pi _\theta}{\partial \theta} + \frac{\partial\pi_\theta}{\partial \mathbf{h}_k}\frac{d \mathbf{h}_k}{d \theta} + \frac{\partial \pi_\theta}{ \partial \mathbf{o}_k} \frac{\partial \mathbf{o}_k}{\partial \mathbf{x}_k}\frac{d \mathbf{x}_k}{ d \theta},
\]
which includes an observation-feedback term at the end, and the same holds for $d \mathbf{h}_k/ d \theta$. Let $\text{sg}(\cdot)$ denote a stop-gradient operator that blocks gradients. The quasi-analytical policy gradient is the gradient of the actor loss \eqref{eq:actor_loss} in which every observation is constant, $\mathbf{o}_k \leftarrow \text{sg}(\mathbf{o}_k)$, removing the feedback term from both $d \mathbf{u}_k/d \theta $ and $d \mathbf{h}_k/ d \theta$, defined as follows:
\begin{equation}
    \label{eq:policy_gradient}
    \tilde{\nabla}_\theta \mathcal{L} = \sum_{k = k_0}^{k_0+h-1} \left(\frac{d \mathbf{u}_k}{d \theta } \bigg|_{\text{sg}(\mathbf{o}_k)}\right)^\top \left( \frac{\partial \mathcal{L}_\theta}{\partial \mathbf{u}_k} \right)^\top,
\end{equation}
where the factor $\partial \mathcal{L}_\theta / \partial \mathbf{u}_k$ is accumulated through the simulator by the adjoint recursion, starting with the terminal adjoint, expressed as follows:
\begin{equation}
    \label{eq:terminal_adjoint}
    \frac{\partial \mathcal{L}_\theta}{\partial \mathbf{x}_{k_0+h}} = -\frac{\gamma^h}{h} \frac{\partial V_\phi}{\partial \mathbf{c}_{k_0+h}}\frac{\partial \Psi}{ \partial \mathbf{x}_{k_0 + h}},
\end{equation}
then compute the state and action adjoints backward to $k_0$:
\begin{equation}
    \begin{aligned}
        \frac{\partial \mathcal{L}_\theta}{\partial \mathbf{x}_k} &= - \frac{\gamma^{k-k_0}}{h} \frac{\partial \mathcal{R}}{\partial \mathbf{x}_k} + \frac{\partial \mathcal{L}_\theta}{\partial \mathbf{x}_{k+1}} \frac{\partial f}{\partial \mathbf{x}_k}, \\
        \frac{\partial \mathcal{L}_\theta}{\partial \mathbf{u}_k} &= -\frac{\gamma^{k-k_0}}{h} \frac{\partial \mathcal{R}}{\partial \mathbf{u}_k} + \frac{\partial \mathcal{L}_\theta}{\partial \mathbf{x}_{k+1}} \frac{\partial f}{\partial \mathbf{u}_k}.
    \end{aligned}
\end{equation}
This recursion incorporates the dynamics Jacobian; thus, $\tilde{\nabla}_\theta{\mathcal{L}}$ is a first-order policy gradient, discarding only the observation-feedback term. From \eqref{eq:terminal_adjoint}, everything beyond the short-horizon $h$ contributes to the policy parameter $\theta$ only through $\partial V_\phi / \partial \mathbf{c}$. The critic parameter $\phi$ is then updated via TD-$\lambda$ after the policy parameter $\theta$ is updated.


\subsection{Two-Stage Training with Warm-Started Value Prior}
We propose a two-stage approach, depicted in Fig. \ref{fig:method}: (i) an expert policy $\pi_\theta^\text{e}$ is trained with access to the gate geometry and the quadrotor's states, and (ii) a vision-based policy $\pi_\theta^\text{v}$, where visual inputs are binary masks of the gate from two ego-centric cameras, is trained in an asymmetric actor-critic manner, warm-starting only the expert’s privileged critic $V_\phi$. Transferring the learnable critic gives the policy a task-shaped terminal adjoint from the first policy update, resulting in better visual policy performance than training the visual actor from scratch or freezing the critic.
To achieve the task, a high-level reward function is defined as:
\begin{equation}
    \label{eq:total_reward}
    \begin{aligned}
        R(\mathbf{x}, \mathbf{u}) = & \ w_\text{prog}r_{\text{prog}} +  w_\text{align}r_{\text{align}} + w_{\text{margin}}r_\text{margin}  \\
        & +w_\text{hover}r_{\text{hover}} + w_\text{action}r_\text{action}.
    \end{aligned}
\end{equation}
The time indices are omitted for brevity. $w$ with a subscript indicates the weights, and the terms are:
\begin{itemize}[leftmargin=1.25em]
    \item Progress reward $r_{\text{prog}}$: the inner product between the agent's velocity and the target direction, i.e., $\mathbf{v}^\top \hat{\mathbf{u}}_{\text{target}}$, where $\hat{\mathbf{u}}_{\text{target}}$ is the unit vector pointing from the agent to the target. The target is initially the gate centroid and switches to the goal upon the agent traversing the gap.
    
    \item Alignment reward $r_{\text{align}}$: a penalty for the agent's roll and pitch misalignment with respect to the gate. Let $\hat{\mathbf{z}}^g = R^\top(\mathbf{q}_g)R(\mathbf{q})\hat{\mathbf{z}}$ denote the agent's body $z$-axis expressed in the gate frame, where $\mathbf{q}_g$ is the gate quaternion. The alignment reward is defined as
    \[
    \label{eq:r_align}
    r_{\text{align}}
    =
    -\| \hat{\mathbf{z}}^g_{[0:1]} \|_2^2
    \exp\left(-\left(\mathbf{p}_{[0]}^g/\sigma_x\right)^2\right) \text{sg}(1-\mathcal{P}),
    \]
    where $\hat{\mathbf{z}}^g_{[0:1]}$ denotes the first two components of $\hat{\mathbf{z}}^g$, $\mathbf{p}^g$ denotes the agent's position in the gate frame, and $\mathbf{p}^g_{[0]}$ is its first component. $\sigma_x$ determines the spread of the Gaussian-like function. $\mathcal{P}$ is an indicator function that returns 1 when the agent has already passed the gap aperture regardless of collision, while it returns 0 otherwise. This term encourages the agent to match its attitude with the gate during traversal.
    \item Clearance shaping reward $r_{\text{margin}}$: a term that rewards the clearance of each corner and penalizes when the corners of $\mathcal{C}$ protrude outside the aperture. Let $\mathbf{c}^g_j \in \mathbb{R}^3$ be the position of the box's $j^\text{th}$ corner in the gate frame, and let $w_a$ and $h_a$ be half of the aperture's width and height in meters, respectively. The term is expressed as:
    \begin{equation}
        \label{eq:r_corner}
        r_\text{margin} = \sum_{j = 0}^{M} \tanh \left( m_j / \sigma_m \right)\exp\left(-\left(\mathbf{c}_{j,[0]}^g/\sigma_x\right)^2\right),
    \end{equation}
    where $m_j = \min (w_a - |\mathbf{c}_{j,[1]}^g|, h_a - |\mathbf{c}_{j, [2]}^g|)$, representing the distance between corner $j$ to the nearest aperture's edge, $M$ is the number of $\mathcal{C}$'s corners, and $\sigma_m = 0.5 h_a$.
    \item Hovering reward $r_\text{hover}$: a term that encourages the agent to approach the goal and hover stably, defined as:
    \begin{equation}
        \label{eq:r_hover}
        r_\text{hover} = \exp\left( -\left({d_\text{goal}/\sigma_p}\right)^2 - \left( {\| \mathbf{v} \|_2/ \sigma_v} \right)^2 \right) \text{sg}(\mathcal{P}),
    \end{equation}
    where $d_\text{goal}$ is an agent's distance-to-goal in meters. $\sigma_p$ and $\sigma_v$ characterize the spread of both terms.
    \item Action regularization term $r_\text{action}$ penalizes actions and action variation to promote energy-efficient trajectories and prevent actuator degradation, expressed as:
    \begin{equation}
        r_{\text{action},k} = - \| \mathbf{u}_k \|_2^2 - \| \mathbf{u}_k - \mathbf{u} _{k-1}\|_2^2.
    \end{equation}
\end{itemize}









\subsection{Simplified Jacobian}

\begin{figure}
    \vspace*{2mm}
    \centering
    \includegraphics[width=1\linewidth]{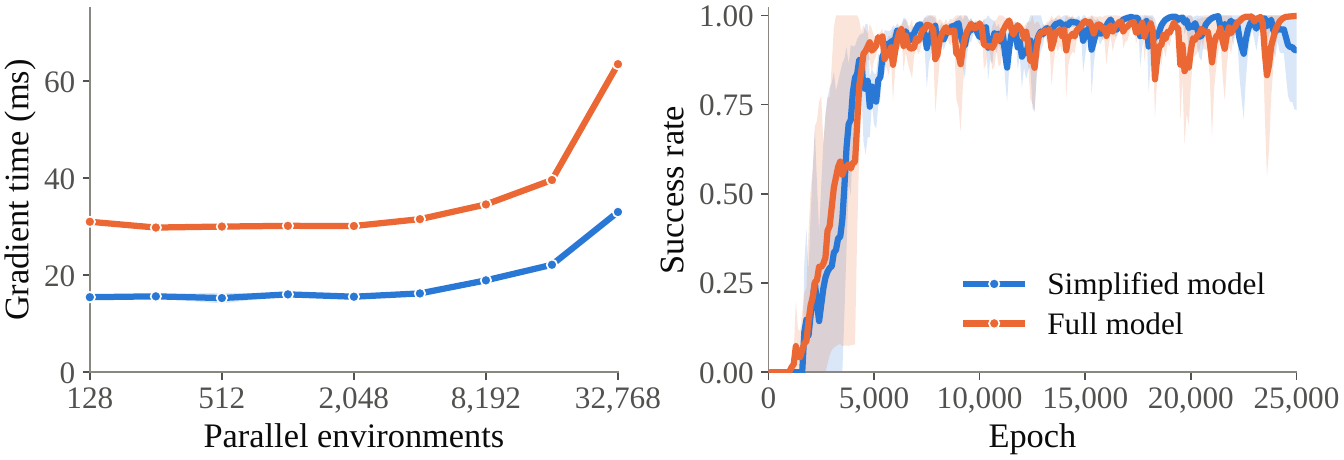}
    \caption{Comparing the gradient computation time (left) and success rate (right; mean and variation over three seeds) between using a simplified or full model to compute gradients.}
    \label{fig:grad_cost}
    \vspace{-0.4cm}
\end{figure}

The presented model includes inner-loop dynamics, which produce stiff gradients and could destabilize training. Additionally, the computation adds $N$ physics substeps per control step, making the gradient computation expensive. We hence employ a simplified surrogate model $f^{\text{sim}}: (\mathbf{x}_k,\mathbf{u}_k ) \mapsto \mathbf{x}_{k+1}^\text{sim}$ for backpropagation, assuming instantaneous body rate tracking and omitting thrust lag, i.e., $(T_k, \boldsymbol{\omega}_k) = (T_k^{\text{ref}}, \boldsymbol{\omega}_k^{\text{ref}})$. The gradient is computed through the simple model via the semi-implicit Euler method without physics substeps at $\Delta t_c$. To make the forward pass value agree with the accurate state, we perform state alignment as follows:
\begin{equation}
    \label{eq:state_alignment}
    \mathbf{x}_{k+1} = \mathbf{x}_{k+1}^\text{sim} + \text{sg}(\mathbf{x}_{k+1}^\text{full} - \mathbf{x}_{k+1}^\text{sim}),
\end{equation}
where $\mathbf{x}^\text{full}_{k+1}$ is the state produced by the accurate model $f$ that includes substeps and inner-loop dynamics. Only factor $\partial f^\text{sim}/ \partial (\mathbf{x}_k, \mathbf{u}_k)$ exists during the gradient computation.
This ensures the rollout value remains valid while backpropagation runs through the simplified dynamics, inspired by \cite{heeg2025learning} and \cite{song2024learning}. Fig. \ref{fig:grad_cost} shows one-batch gradient computation time on NVIDIA GeForce RTX 5080 and training curves of policy learning via QPG, using a simplified and full model to compute gradients. The figure shows that using the simplified model enhances training without sacrificing performance.

\subsection{Visual Observation Renderer}
\label{sec:renderer}

During training and evaluation, we utilize the gate's binary mask as a visual observation, obtained through perspective transformation and rasterization at every control step. We use a mask rather than a photorealistic or depth image to simplify training and experimentation, as it retains the gap geometry, bypasses texture, lighting, and motion blur in the sim-to-real gap, and is analytic; thus, it can be compiled and vectorized with the simulator. Let $\mathbf{g}_{n}^{c}$ be the $n^{\text{th}}$ gate corner position in the camera frame, where the optical axis is along $\mathbf{g}_{n,[0]}^{c}$. $\mathbf{a}_n = [u_n, v_n]^\top$ is the gate corner in pixel coordinates, computed through pinhole projection. The aperture's edges are constructed cyclically via $\mathbf{l}_n(t_n) = \mathbf{a}_{n}+t_n\mathbf{d}_n$, where $\mathbf{d}_n  = \mathbf{a}_{n+1}-\mathbf{a}_n, \ \forall n=0,1, \dots, L-1$, and $L$ is the number of aperture's corner.
For a pixel center $\mathbf{b} = [u,v]^\top$, the squared distance from the pixel to the edge is
\begin{equation}
    \label{eq:distance_to_segment}
    d_n^2(\mathbf{b}) = \min_{s \in [0,1]} \| \mathbf{b} - \mathbf{l}_n (s) \| _2^2,
\end{equation}
and the gate's aperture mask is defined as:
\begin{equation}
    \label{eq:mask}
    m_{\text{mask}} (\mathbf{b}) = \mathds{1}[ \min_{n \in \mathcal{E}} d_n^2(\mathbf{b}) < (w_l/2)^2],
\end{equation}
where $\mathds{1}[\cdot]$ is an indicator function, $w_l$ is a line width, and $\mathcal{E}$ is an index set of edges for which at least one endpoint is in front of the camera. When traversing the gap, a corner may be behind the camera, and the drawn line is incorrect. To avoid this, we adjust the edge endpoint by sliding it along the edge to the point where it crosses the $\mathbf{g}_{n, [0]}^c = 0.001$ plane and project this point instead. This process repeats for the second camera, and the policy's visual observation is the concatenation of the two binary masks, shown in Fig. \ref{fig:method} right.


\subsection{Training Details}

We implement the trainer and environment using JAX \cite{jax2018github}.
The episode length is $500$ steps at a control frequency of $100$ Hz, and the short-horizon length $h$ is $64$. For both stages, the policy and critic initial learning rates are $5 \times 10^{-4}$ and $2\times 10^{-3}$, respectively, and decay linearly. The policy is optimized using Adam via Optax for $25,000$ iterations. The discount factor $\gamma$ is $0.99$ for both stages. The critic is learned via TD-$\lambda$ with $\lambda=0.95$. We train the policies with $1024$ parallel environments on NVIDIA A40 GPUs. The weights for each term in the total one-step reward function \eqref{eq:total_reward} are as follows: $w_\text{prog} = 1.0$, $w_{\text{align}} = 8.0$,  $w_\text{margin}=12.0$, $w_\text{hover} = 10.0$, and $w_{\text{action}} = 0.01$. The parameters in reward terms \eqref{eq:r_corner} and \eqref{eq:r_hover} are $\sigma_x = 0.6$, $\sigma_p=0.25$, and $\sigma_v= 0.5$. The quadrotor parameters are specified in Table \ref{tab:ft_plant} in Nominal column.  For the gate renderer parameter, $w_l = 1.5$ pixels.
\section{Experiments}


This section evaluates the key components of our approach. First, QPG is used to pretrain a task-informed privileged critic, then its value function is transferred to warm-start visual policy training. We compare our method with baseline approaches, show ablation studies, and evaluate generalization to unseen gap shapes. We further compare it with the two-stage reinforcement–imitation learning approach similar to the prior works and validate the trained visual policy in the real world. All reported results are based on three random seeds.




\subsection{Pretraining a Task-Informed Critic}
\label{sec:expert}

\begin{figure}
    \vspace*{2mm}
    \centering
    \includegraphics[width=0.9\linewidth]{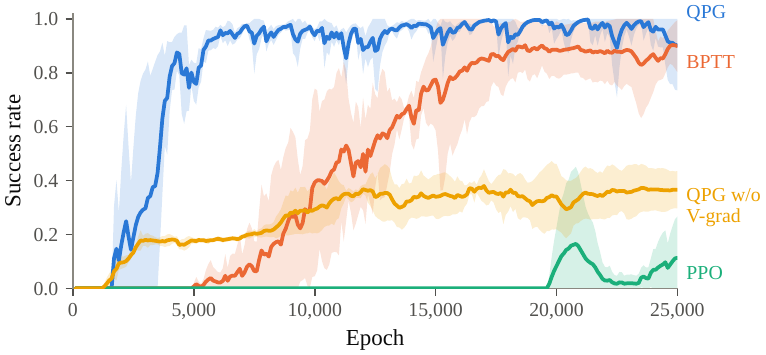}
    \caption{Training curves of the privileged pretraining stage: QPG, full-episode BPTT, PPO, and the ablation without the critic gradient (QPG w/o V-grad).}
    \label{fig:state_learning_curve}
    \vspace{-0.4cm}
\end{figure}

In this stage, both networks receive the same privileged observations, including body-frame velocity $\mathbf{v}^b$, attitude $\mathbf{q}$, accelerometer readings $\tilde{\mathbf{a}}^b$, the four gate-corner positions in body frame $\mathbf{g}^b_{n}$, the goal position in body frame $\mathbf{p}^b_{\text{goal}}$, and the previous action $\mathbf{u}_{k-1}$. The networks are modeled as three-layer MLPs with 256 units each and ELU activations. We compare the expert policy trained using our approach with the PPO and BPTT baselines. The latter baseline refers to untruncated full-episode BPTT without the terminal-value bootstrap. We also ablate the value gradient by detaching the terminal value, which is equivalent to $h$-step BPTT. This isolates the value-gradient contribution at a fixed horizon. In Fig. \ref{fig:state_learning_curve}, QPG (blue) reaches 0.8 in 12 minutes and saturates near 1.0, while BPTT (orange) reaches a success rate close to that of QPG, but it requires 16000 epochs. Detaching the terminal value (QPG w/o V-grad; yellow) rises earlier, but then saturates at 0.38. Finally, the PPO curve (green) remains at zero for the first 19,500 epochs in our settings, consistent with the result reported in \cite{wu2025whole} during the first-stage RL training without the informed reset.

From the results, PPO has to infer the gradients from sampled discounted returns. Without directed exploration, the rollouts almost never attain the tight success set. BPTT and QPG, instead, provide more useful gradients by differentiating a task loss through differentiable physics, resulting in more effective policy updates. However, BPTT propagates through 500-step dynamics Jacobians, which can cause exploding gradients and unstable training. Although it still achieves a high success rate, it does so more slowly and requires more samples than QPG, which backpropagates through only 64 steps and utilizes the critic’s gradient to approximate long-term credit assignment. Without the value gradient (w/o V-grad), the policy update accounts for only 64 steps ahead, and the curve plateaus.

\subsection{Vision-based Policy Learning}
\label{sec:vision_policy}

\begin{figure}
    \vspace*{2mm}
    \centering
    \includegraphics[width=1\linewidth]{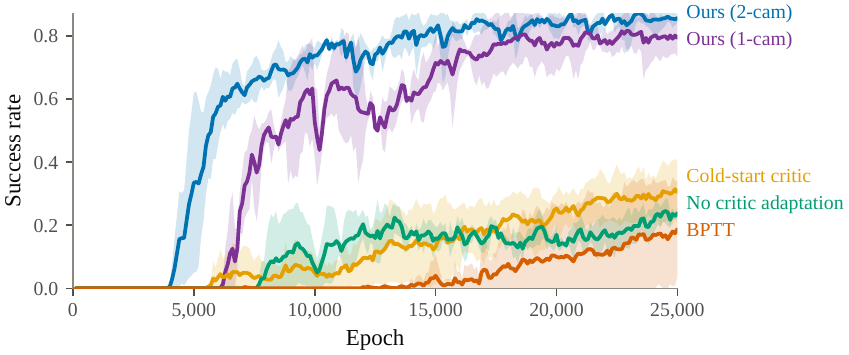}
    \caption{Training curves of the second stage: ours with two- or one-camera configuration, full-episode BPTT, and the ablations with cold-start critic and a frozen critic (no critic adaptation).}
    \label{fig:succ_vision}
    \vspace{-0.4cm}
\end{figure}

The visual policy accepts the rendered binary mask $I_k$ (Section \ref{sec:renderer}), augmented with low-dimensional observations: the agent's roll $\phi$, pitch $\theta$, an accelerometer reading $\tilde{\mathbf{a}}^b$, a goal position in the body frame $\mathbf{p}^b_\text{goal}$, and the previous action $\mathbf{u}_{k-1}$. The policy architecture consists of a convolutional neural network (CNN), a gated recurrent unit (GRU), and a two-layer MLP. The $84 \times 300$ concatenated mask, $84 \times 150$ from each of the two cameras, passes through the CNN (32/64/64 filters, 8/4/3 kernels, 4/2/1 strides); the convolution kernel slides across both views, allowing the network to fuse cross-camera information early. The flattened output is then concatenated with the low-dimensional observations (scaled by a fixed gain of 10) and is passed to a 256-unit GRU that carries temporal context across the rollout, followed by a two-layer MLP.
The camera configuration is similar to that in \cite{liu2024omninxt}, except that only frontal cameras are used. We use the same extrinsics as in our real platform, where both camera centers are $10$ cm forward, $2.9$ cm up, and $\pm4.2$ cm left and right relative to the agent centroid. The optical axes splay $\pm45^\circ$ outward from the quadrotor front. The fisheye lenses are approximated with a field-of-view-matched pinhole model rather than replicating their distortion.
The training curves of this stage are shown in Fig. \ref{fig:succ_vision}, where we train the policy via QPG, warm-starting the value function from the pretrained critic in the first stage (blue), and compare the policy trained using our approach with full-episode BPTT without a value function (orange). We also ablate our approach with three scenarios: (i) training both critic and policy from scratch using QPG (cold-start critic; yellow), (ii) freezing the critic from the first stage while training the visual policy (no critic adaptation; green), and (iii) changing the visual observation to a single face-forward camera (purple).

\begin{figure}
    \centering
    \includegraphics[width=1\linewidth]{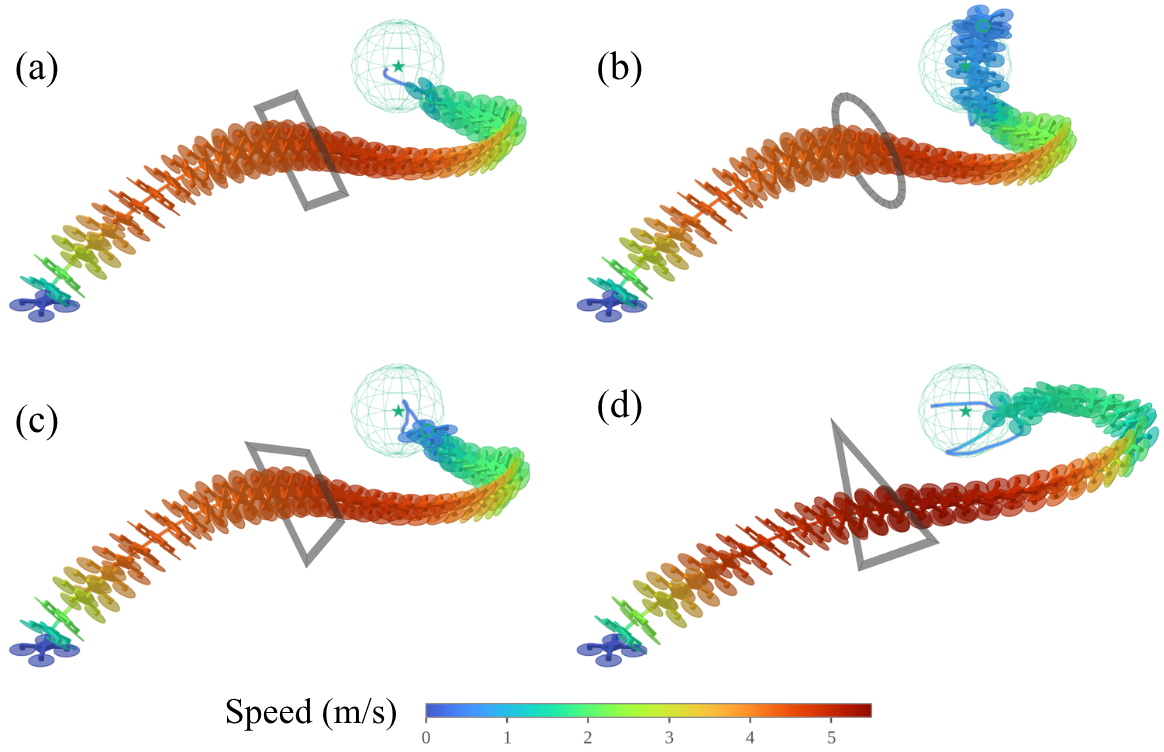}
    \caption{Trajectories of a quadrotor gap traversal using our vision-based policy for different gap shapes. (a) rectangular gap, similar to the one used during training, (b) elliptical gap, (c) trapezoidal gap, and (d) triangular gap. The spherical shapes in each image represent the goal region.}
    \label{fig:demonstrations}
    \vspace{-0.4cm}
\end{figure}

The results show that our trained visual policy achieves a peak success rate of 0.893, while BPTT, although its training curve grows steadily, requires more samples as it cannot leverage privileged information from the critic. In the ablation study, the cold-start critic with QPG outperforms BPTT as it implicitly utilizes privileged information from the asymmetric actor-critic structure and the training is more stable, but it still cannot match the two-stage approach, showing that the two-stage approach substantially improves learning. For the frozen critic, the pretrained critic provides useful initial guidance but remains below 0.25 afterward, as its value function is fixed to the expert's state distribution and cannot adapt to the visual policy's. The single-camera ablation shows that the two-outward-splayed camera setting improves performance slightly due to its wider field of view, which keeps the gate in view as the agent passes it.
Fig. \ref{fig:demonstrations} (a) displays the trajectory of the agent executing the trained policy (our approach) to pass the rectangular gap and hover stably at the goal with a maximum speed of up to 5.0 m/s.
Interestingly, we demonstrate that our trained vision-based policy traverses gap shapes never exploited during training. We replace the rectangular gap with an elliptical, trapezoidal, and triangular gap while using the same policy trained only on the rectangular gap. The same policy completes each traversal with unseen shapes in Fig. \ref{fig:demonstrations} (b)--(d).

\textit{Why warm-starting matters}:
For the rollout step beyond 64 short-horizon, credit reaches the policy only through the terminal adjoint. Also, with detached observations, the critic provides a feedback pathway to the policy gradients. Thus, initially, a cold-started critic supplies an arbitrary direction while the policy is still learning. In the two-stage approach, the pretrained critic provides the actor with a task-shaped boundary derivative from the first epoch, which serves as adaptable and useful guidance. The same critic can also warm-start an actor with different observations.

\subsection{Transfer Under a Plant Revision}
\label{subsec:plant_revise}

\begin{figure}
    \centering
    \includegraphics[width=0.9\linewidth]{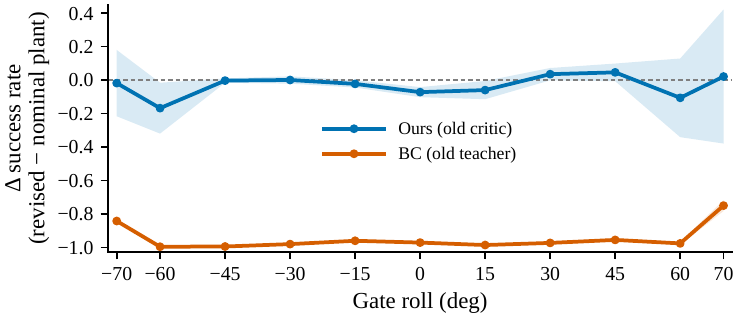}
    \caption{Success rate differences between training with the revised plant (with the old critic or teacher) and the nominal plant. Our two-stage approach is shown in blue, and behavior cloning (BC) is shown in orange.}
    \label{fig:plant_fine_tuning_delta}
    \vspace{-0.4cm}
\end{figure}

A practical advantage of our approach is that it does not require retraining the expert policy when plant parameters change, since only the second stage is rerun with the old pretrained critic to train the policy for the revised quadrotor plant. We construct a revised plant by changing parameters of the nominal plant as shown in Table \ref{tab:ft_plant}. We employ data aggregation (DAgger) \cite{ross2011reduction} to perform behavior cloning, where the experience is generated with the revised plant while action labels are queried from the old expert.
Fig. \ref{fig:plant_fine_tuning_delta} shows the difference in success rates between the policy trained on the revised and the nominal plants. For behavior cloning (orange), although it yields a strong baseline when the plant parameters match those used to train the expert policy, it shows significant degradation when distilling from the old teacher. On the other hand, our two-stage approach (blue) can reuse the old pretrained critic to train the second stage for the revised dynamics without sacrificing performance.



\begin{table}[t]
  \vspace*{2mm}
  \centering
  \caption{Plant parameters of the nominal and revised quadrotor.}
  \label{tab:ft_plant}
  \begin{tabular}{lccc}
  \toprule
  Parameter & Nominal & Revised  \\
  \midrule
  Mass $m$ [kg]                & 1.279 & 1.407 ($+10\%$) \\
  Body-rate's $\omega_n$ [rad/s] & 100  & 70 \\
  Body-rate damping $\zeta$             & 1.00  & 0.80\\
  Body-rate DC gain $K$                 & 0.58  & 0.75\\
  Thrust time constant $\tau$ [ms]      & 12.5  & 25 \\
  \bottomrule
  \end{tabular}
\end{table}

\subsection{Real-world Experiments}
\label{sec:real}

\begin{figure} [ht]
    \centering
    \includegraphics[width=1\linewidth]{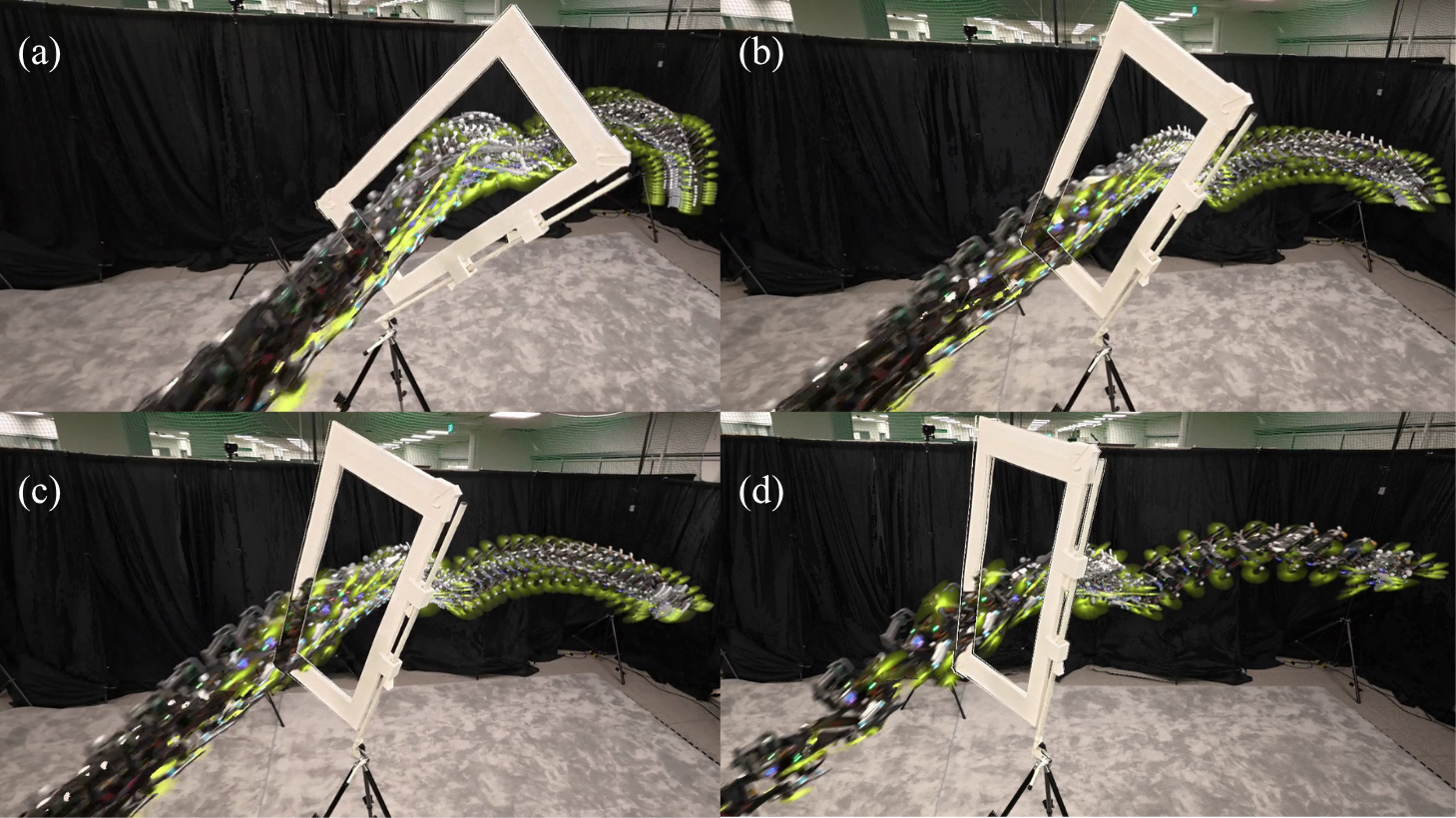}
    \caption{Time-lapse images of the real-world experiments. The trained visual policy is executed with a gate binary mask observation rendered online. The gate roll angle varies as follows: (a) $45^\circ$, (b) $60^\circ$, (c) $70^\circ$, and (d) $80^\circ$.}
    \label{fig:real}
    \vspace{-0.4cm}
\end{figure}

We deploy our visual policy on a $1.279$ kg quadrotor platform with the dimensions shown in Fig. \ref{fig:dimensions} left. The onboard computer is an NVIDIA Jetson Orin NX, connected to a Holybro Pixhawk 6X Mini flight controller, operating the PX4 autopilot \cite{meier2015px4}. We synthesize a binary mask for the visual observation with the mechanism in \ref{sec:renderer} and the configurations explained in \ref{sec:vision_policy}.
In the experiment, a motion capture system provides the poses of the gate and the quadrotor, from which the gate's binary mask is rendered online and additional low-dimensional observations are computed. The experiment is conducted in a highly confined space with a $4 \times 4$ $\text{m}^2$ area and a $2$ m operating height. The trained policy is exported to ONNX and runs via its Runtime C++ on Jetson.
We construct a real rectangular gate with the dimensions shown in Fig. \ref{fig:dimensions} (right) for the real quadrotor to perform the task. We conduct four experiments at four gate roll angles: $45^\circ, \ 60^\circ, \ 70^\circ,$ and $80^\circ$. The real-world experiments in a confined space are shown in Fig. \ref{fig:real} with an average maximum speed of $3.5$ m/s.

\section{Conclusion}


We propose a two-stage policy learning, leveraging differentiable simulation for a quadrotor agile gap traversal. The visual policy observes a binary gate mask from two ego-centric cameras plus extra low-dimensional observation, without knowing the gap geometry. We show that warm-starting a privileged critic trained with an expert policy greatly improves visual policy training. The trained policy also traverses through unseen gap shapes. We show the advantage of our approach over the previous RL-imitation learning method when the plant parameters change, as ours does not require retraining the expert policy. Furthermore, we deploy the trained visual policy on a real quadrotor.

For limitations and future work, in this experiment, the binary gate mask is rendered from motion-capture poses rather than obtained from the onboard cameras, neglecting perception flaws. Thus, a robust binary gate predictor \cite{Geles-RSS-24} or an end-to-end policy from raw images specifically for tight-gap traversal would eliminate the need to synthesize the mask online. Moreover, the policy receives the goal's relative position; following \cite{liu2024omninxt}'s camera configuration, it could learn to stabilize behind the gate after traversing without it. Finally, the proposed scheme could also be applied to other challenging robotics applications.



\bibliographystyle{IEEEtran}
\bibliography{references}

@article{chou1992quaternion,
  title={Quaternion kinematic and dynamic differential equations},
  author={Chou, Jack CK},
  journal={IEEE Transactions on robotics and automation},
  volume={8},
  number={1},
  pages={53--64},
  year={1992},
  publisher={IEEE}
}

@article{chen2022learning,
  title={Learning real-time dynamic responsive gap-traversing policy for quadrotors with safety-aware exploration},
  author={Chen, Shiyu and Li, Yanjie and Lou, Yunjiang and Lin, Ke and Wu, Xinyu},
  journal={IEEE Transactions on Intelligent Vehicles},
  volume={8},
  number={3},
  pages={2271--2284},
  year={2022},
  publisher={IEEE}
}

@article{freeman2021brax,
  title={Brax--a differentiable physics engine for large scale rigid body simulation},
  author={Freeman, C Daniel and Frey, Erik and Raichuk, Anton and Girgin, Sertan and Mordatch, Igor and Bachem, Olivier},
  year={2021},
  journal={Advances in Neural Information Processing Systems}
}

@inproceedings{falanga2017aggressive,
  title={Aggressive quadrotor flight through narrow gaps with onboard sensing and computing using active vision},
  author={Falanga, Davide and Mueggler, Elias and Faessler, Matthias and Scaramuzza, Davide},
  booktitle={2017 IEEE international conference on robotics and automation (ICRA)},
  pages={5774--5781},
  year={2017},
  organization={IEEE}
}

@INPROCEEDINGS{Geles-RSS-24, 
    AUTHOR    = {Ismail Geles AND Leonard Bauersfeld AND Angel Romero AND Jiaxu Xing AND Davide Scaramuzza}, 
    TITLE     = {{Demonstrating Agile Flight from Pixels without State Estimation}}, 
    BOOKTITLE = {Proceedings of Robotics: Science and Systems}, 
    YEAR      = {2024}, 
    ADDRESS   = {Delft, Netherlands}, 
    MONTH     = {July}, 
    DOI       = {10.15607/RSS.2024.XX.082} 
}

@article{hu2025narrow,
  title={Narrow Gap Traversing via Differentiable Physics},
  author={Hu, Yu and Zou, Danping},
  journal={Journal of Shanghai Jiaotong University (Science)},
  pages={1--8},
  year={2025},
  publisher={Springer}
}

@inproceedings{haarnoja2018soft,
  title={Soft actor-critic: Off-policy maximum entropy deep reinforcement learning with a stochastic actor},
  author={Haarnoja, Tuomas and Zhou, Aurick and Abbeel, Pieter and Levine, Sergey},
  booktitle={International conference on machine learning},
  pages={1861--1870},
  year={2018},
  organization={PMLR}
}

@article{hu2025seeing,
  title={Seeing through pixel motion: Learning obstacle avoidance from optical flow with one camera},
  author={Hu, Yu and Zhang, Yuang and Song, Yunlong and Deng, Yang and Yu, Feng and Zhang, Linzuo and Lin, Weiyao and Zou, Danping and Yu, Wenxian},
  journal={IEEE Robotics and Automation Letters},
  year={2025},
  publisher={IEEE}
}

@inproceedings{heeg2025learning,
  title={Learning quadrotor control from visual features using differentiable simulation},
  author={Heeg, Johannes and Song, Yunlong and Scaramuzza, Davide},
  booktitle={2025 IEEE International Conference on Robotics and Automation (ICRA)},
  pages={4033--4039},
  year={2025},
  organization={IEEE}
}

@software{jax2018github,
  author = {James Bradbury and Roy Frostig and Peter Hawkins and Matthew James Johnson and Yash Katariya and Chris Leary and Dougal Maclaurin and George Necula and Adam Paszke and Jake Vander{P}las and Skye Wanderman-{M}ilne and Qiao Zhang},
  title = {{JAX}: composable transformations of {P}ython+{N}um{P}y programs},
  url = {http://github.com/jax-ml/jax},
  version = {0.3.13},
  year = {2018},
}

@inproceedings{liu2024omninxt,
  title={Omninxt: A fully open-source and compact aerial robot with omnidirectional visual perception},
  author={Liu, Peize and Feng, Chen and Xu, Yang and Ning, Yan and Xu, Hao and Shen, Shaojie},
  booktitle={2024 IEEE/RSJ International Conference on Intelligent Robots and Systems (IROS)},
  pages={10605--10612},
  year={2024},
  organization={IEEE}
}

@inproceedings{lin2019flying,
  title={Flying through a narrow gap using neural network: an end-to-end planning and control approach},
  author={Lin, Jiarong and Wang, Luqi and Gao, Fei and Shen, Shaojie and Zhang, Fu},
  booktitle={2019 IEEE/RSJ international conference on intelligent robots and systems (IROS)},
  pages={3526--3533},
  year={2019},
  organization={IEEE}
}

@inproceedings{lee2010geometric,
  title={Geometric tracking control of a quadrotor UAV on SE (3)},
  author={Lee, Taeyoung and Leok, Melvin and McClamroch, N Harris},
  booktitle={49th IEEE conference on decision and control (CDC)},
  pages={5420--5425},
  year={2010},
  organization={IEEE}
}

@article{liu2018search,
  title={Search-based motion planning for aggressive flight in SE (3)},
  author={Liu, Sikang and Mohta, Kartik and Atanasov, Nikolay and Kumar, Vijay},
  journal={IEEE Robotics and Automation Letters},
  volume={3},
  number={3},
  pages={2439--2446},
  year={2018},
  publisher={IEEE}
}

@inproceedings{meier2015px4,
  title={PX4: A node-based multithreaded open source robotics framework for deeply embedded platforms},
  author={Meier, Lorenz and Honegger, Dominik and Pollefeys, Marc},
  booktitle={2015 IEEE international conference on robotics and automation (ICRA)},
  pages={6235--6240},
  year={2015},
  organization={IEEE}
}

@inproceedings{mellinger2011minimum,
  title={Minimum snap trajectory generation and control for quadrotors},
  author={Mellinger, Daniel and Kumar, Vijay},
  booktitle={2011 IEEE international conference on robotics and automation},
  pages={2520--2525},
  year={2011},
  organization={Ieee}
}

@inproceedings{pinto2018asymmetric,
  author    = {Pinto, Lerrel and Andrychowicz, Marcin and Welinder, Peter and Zaremba, Wojciech and Abbeel, Pieter},
  title     = {Asymmetric Actor Critic for Image-Based Robot Learning},
  booktitle = {Proceedings of Robotics: Science and Systems},
  year      = {2018},
  address   = {Pittsburgh, Pennsylvania},
  month     = {June},
  doi       = {10.15607/RSS.2018.XIV.008}
}

@inproceedings{ross2011reduction,
  title={A reduction of imitation learning and structured prediction to no-regret online learning},
  author={Ross, St{\'e}phane and Gordon, Geoffrey and Bagnell, Drew},
  booktitle={Proceedings of the fourteenth international conference on artificial intelligence and statistics},
  pages={627--635},
  year={2011},
  organization={JMLR Workshop and Conference Proceedings}
}

@article{romero2025actor,
  title={Actor--critic model predictive control: Differentiable optimization meets reinforcement learning for agile flight},
  author={Romero, Angel and Aljalbout, Elie and Song, Yunlong and Scaramuzza, Davide},
  journal={IEEE Transactions on Robotics},
  volume={42},
  pages={673--692},
  year={2025},
  publisher={IEEE}
}

@article{sun2026learning,
  title={Learning Agile Gate Traversal via Analytical Optimal Policy Gradient},
  author={Sun, Tianchen and Wang, Bingheng and Gerdpratoom, Nuthasith and Tang, Longbin and Gao, Yichao and Zhao, Lin},
  journal={arXiv preprint arXiv:2508.21592},
  year={2026},
  note={Accepted to 2026 IEEE/RSJ International Conference on Intelligent Robots and Systems (IROS)}
}

@article{su2026vector,
  title={Vector Field Augmented Differentiable Policy Learning for Vision-Based Drone Racing},
  author={Su, Yang and Yu, Feng and Hu, Yu and Niu, Xinze and Zhang, Linzuo and Sun, Fangyu and Zou, Danping},
  journal={IEEE Robotics and Automation Letters},
  year={2026},
  publisher={IEEE}
}

@InProceedings{song2024learning,
  title = 	 {Learning Quadruped Locomotion Using Differentiable Simulation},
  author =       {Song, Yunlong and Kim, Sangbae and Scaramuzza, Davide},
  booktitle = 	 {Proceedings of The 8th Conference on Robot Learning},
  pages = 	 {258--271},
  year = 	 {2025},
  volume = 	 {270},
  series = 	 {Proceedings of Machine Learning Research},
  publisher =    {PMLR},
}

@inproceedings{schulman2015trust,
  title={Trust region policy optimization},
  author={Schulman, John and Levine, Sergey and Abbeel, Pieter and Jordan, Michael and Moritz, Philipp},
  booktitle={International conference on machine learning},
  pages={1889--1897},
  year={2015},
  organization={PMLR}
}

@article{schulman2017proximal,
  title={Proximal policy optimization algorithms},
  author={Schulman, John and Wolski, Filip and Dhariwal, Prafulla and Radford, Alec and Klimov, Oleg},
  journal={arXiv preprint arXiv:1707.06347},
  year={2017}
}

@article{xie2023learning,
  title={Learning agile flights through narrow gaps with varying angles using onboard sensing},
  author={Xie, Yuhan and Lu, Minghao and Peng, Rui and Lu, Peng},
  journal={IEEE Robotics and Automation Letters},
  volume={8},
  number={9},
  pages={5424--5431},
  year={2023},
  publisher={IEEE}
}

@article{xiao2021flying,
  title={Flying through a narrow gap using end-to-end deep reinforcement learning augmented with curriculum learning and Sim2Real},
  author={Xiao, Chenxi and Lu, Peng and He, Qizhi},
  journal={IEEE transactions on neural networks and learning systems},
  volume={34},
  number={5},
  pages={2701--2708},
  year={2021},
  publisher={IEEE}
}

@inproceedings{xu2022accelerated,
  title={Accelerated Policy Learning with Parallel Differentiable Simulation},
  author={Xu, Jie and Makoviychuk, Viktor and Narang, Yashraj and Ramos, Fabio and Matusik, Wojciech and Garg, Animesh and Macklin, Miles},
  booktitle={International Conference on Learning Representations},
  year={2022}
}

@inproceedings{wang2023learning,
  title={Learning agile flight maneuvers: Deep SE (3) motion planning and control for quadrotors},
  author={Wang, Yixiao and Wang, Bingheng and Zhang, Shenning and Sia, Han Wei and Zhao, Lin},
  booktitle={2023 IEEE International Conference on Robotics and Automation (ICRA)},
  pages={1680--1686},
  year={2023},
  organization={IEEE}
}

@article{wu2026precise,
  title={Precise aggressive aerial maneuvers with sensorimotor policies},
  author={Wu, Tianyue and Xu, Guangtong and Wang, Zihan and Lin, Junxiao and Chen, Tianyang and Wu, Yuze and Han, Zhichao and Liu, Zhiyang and Gao, Fei},
  journal={Science Robotics},
  volume={11},
  number={115},
  pages={eaeb0180},
  year={2026},
  publisher={American Association for the Advancement of Science}
}

@inproceedings{wu2025whole,
  title={Whole-body control through narrow gaps from pixels to action},
  author={Wu, Tianyue and Chen, Yeke and Chen, Tianyang and Zhao, Guangyu and Gao, Fei},
  booktitle={2025 IEEE International Conference on Robotics and Automation (ICRA)},
  pages={11317--11324},
  year={2025},
  organization={IEEE}
}

@inproceedings{wiedemann2023training,
  title={Training Efficient Controllers via Analytic Policy Gradient},
  author={Wiedemann, Nina and W{\"u}est, Valentin and Loquercio, Antonio and M{\"u}ller, Matthias and Floreano, Dario and Scaramuzza, Davide},
  booktitle={2023 IEEE International Conference on Robotics and Automation (ICRA)},
  pages={1349--1356},
  year={2023},
  organization={IEEE}
}

@article{you2026accelerating,
  title={Accelerating visual-policy learning through parallel differentiable simulation},
  author={You, Haoxiang and Liu, Yilang and Abraham, Ian},
  journal={Advances in Neural Information Processing Systems},
  volume={38},
  pages={51896--51925},
  year={2026}
}

@article{zhang2026vision,
  title={Vision-Based End-to-End Learning for UAV Traversal of Irregular Gaps via Differentiable Simulation},
  author={Zhang, Linzuo and Hu, Yu and Yu, Feng and Deng, Yang and Yu, Wenxian and Zou, Danping},
  journal={IEEE Robotics and Automation Letters},
  year={2026},
  publisher={IEEE}
}

@article{zhang2025learning,
  title={Learning vision-based agile flight via differentiable physics},
  author={Zhang, Yuang and Hu, Yu and Song, Yunlong and Zou, Danping and Lin, Weiyao},
  journal={Nature Machine Intelligence},
  volume={7},
  number={6},
  pages={954--966},
  year={2025},
  publisher={Nature Publishing Group UK London}
}

\end{document}